# A Responsible Artificial Intelligence Framework for Groundwater Modeling

Chong Chen[1,*], Yulu Zhang[1], Qingxi Guo[1], Yihan Liu[1]

[1] College of Artificial Intelligence, China University of Petroleum – Beijing

[*] Email: chenchong@cup.edu.cn

**Abstract:** The rapid development and widespread application of artificial intelligence (AI) have sparked intense discussions on how to deploy responsible AI systems in a manner aligned with human values and ethical standards. Compared to fields like healthcare, energy, or finance, the application of AI in groundwater is relatively limited, and research on responsible AI is even more scarce. Taking the middle reaches of the Heihe River Basin as the study area, this paper proposes six Responsible AI principles—transparency, technical robustness, privacy governance, fairness, accountability, and sustainability. LSTM and Transformer time-series models are developed using multi-source hydrometeorological data, and validated via post-hoc interpretability, Monte Carlo simulation, and scenario analysis. The results show that Transformer outperforms LSTM in accuracy, robustness, and interpretability, demonstrating the operability and practical value of Responsible AI principles in groundwater prediction to support sustainable water management under climate change and human activities.

## 1 Introduction

The rapid advancement of artificial intelligence has brought transformative opportunities to computer science, environmental science, medical informatics, finance and many other fields, accelerating the transition of human society from a traditional information society to an AI-driven society centered on artificial intelligence. Nevertheless, as AI continues to advance in data processing capabilities and overall performance, a host of complex ethical challenges and potential security risks have gradually emerged. On the one hand, the "black-box effect" of neural networks undermines the transparency of AI decision-making[1]. With intricate internal parameter mappings, it is difficult to interpret the formation logic of predictions and identify key influencing factors. On the other hand, AI relies heavily on massive datasets for operation, which raises risks of data leakage. Furthermore, Google was once accused of racial bias after its facial recognition system misclassified photos of African-American users[2]. Issues such as algorithmic prejudice and privacy infringement caused by AI remain difficult to address effectively. How to fully tap the potential of AI technology while mitigating the associated governance risks and social uncertainties has become an urgent issue for academic researchers and enterprise managers. Against this background, Responsible Artificial Intelligence (Responsible AI) provides new ideas to address the above issues by standardizing the secure, ethical and trustworthy design, development, evaluation and deployment of AI systems[3]. It prioritizes accountability and controllable social impacts alongside technical performance, aiming to create human-machine collaborative, risk-controllable and value-aligned intelligent systems, and has gained widespread consensus in global AI governance.

In response to these challenges, the international community is gradually reaching a consensus. The European Union (EU) aims to establish human-centric and trustworthy ethical standards for AI. It has pioneered the release of the *Ethics Guidelines for Trustworthy AI* in 2019, seven key requirements for promoting the sound development of Trustworthy AI should be sustained over three main pillars throughout the system's entire life cycle, i.e., lawful, ethical, and robust[4]. The United Arab Emirates issued the *AI Ethics Principles and Guidelines* in 2022, clarifying goals concerning AI infrastructure development and AI regulation. In 2024, Singapore, a regional example of balancing AI innovation and governance in Southeast Asia, launched the forward-looking *Generative AI Governance Framework* for large language models and generative AI, urging responsible development and deployment of reliable AI systems. And the EU introduced the *AI Act*, the first comprehensive legal framework for AI regulation[5], which not only affects the global AI governance pattern but also significantly influences the development path of AI in China. However, as AI exerts growing social and environmental impacts, theoretical frameworks alone can no longer address complex real-world risks. Building on Trustworthy AI, Responsible AI adopts concrete auditing and supervision measures to translate "trustworthiness" into practical accountability. It clarifies accountability for biased or faulty AI models, and upholds fairness, sustainability and public interests throughout the AI lifecycle.

Groundwater is a vital water resource for agricultural irrigation, industrial production and domestic water supply. Driven by global climate change, its recharge, runoff and discharge have undergone dramatic changes. Due to the concealment and spatiotemporal heterogeneity of groundwater systems, traditional hydrological models built on simplified physical assumptions fail to accurately reflect groundwater evolution. By contrast, deep learning can mine massive datasets, capture complex nonlinear relationships and deliver timely, precise predictions, emerging as a key research direction in water resources[6,7]. Given the characteristics of groundwater systems, this paper proposes a Responsible AI governance framework for groundwater models, integrating transparency, technical robustness, fairness and accountability into the full lifecycle of models. Empirical studies based on LSTM and Transformer are conducted. The results provide new technical support for sustainable groundwater management and a reference for AI application in other high-risk fields.

# 2 Research Methods

## 2.1 Responsible AI

"Responsible" refers to the ability to bear the corresponding consequences and be accountable for them when making decisions and taking actions. As AI models and governance frameworks become increasingly diverse, related terms such as Trustworthy AI and Ethical AI have emerged[8,9], reflecting differences in understanding brought by various professional backgrounds and practical perspectives. However, these terms are often used interchangeably and even regarded as synonyms. Of course, the proposal of Responsible AI does not rely solely on researchers; rather, it is a consensus concept promoted jointly by the global academic community, industry, policymakers, and social organizations during the process of AI rapidly permeating various sectors of society. To date, a unified and universally accepted definition has not yet been fully formed. Individual researchers consider Responsible AI as the process of translating Trustworthy AI from theory to practice and regulation, distilling a highly generalized operational definition[10]:

**Definition.** A Responsible AI system is an AI system that requires ensuring auditability and accountability during its design, development and use, according to specifications and the applicable regulation of the domain of practice in which the AI system is to be used.

This definition highlights the need for traceability and regulatory compliance across the entire AI lifecycle. By embedding responsibility into each stage of development and deployment, it helps mitigate bias at its root, enhances model transparency and explainability, and fosters greater trust between AI systems and their users. Researchers must deeply consider how to follow "Responsible" standards in the development and application of artificial intelligence. The AI principles proposed by different entities have different focuses: enterprises pay attention to the practical implementation of technology, while governments and international organizations focus on the standardization of governance frameworks. However, essentially, the core of these principles has certain similarities, which can be summarized into five main principles:

- *Explainability:* express the important factors affecting the results in a way that humans can understand.
- *Safety:* have the ability to be free from safety risks, risks that are intolerable.
- *Accountability:* establish mechanisms to ensure responsibility and accountability for artificial intelligence systems and their outcomes.
- *Fairness:* unfair outputs caused by AI systems should be reduced. In other words, AI systems ought to serve all individuals and remain free from bias.
- *Privacy:* implementation and maintenance of confidentiality of data and system attributes.

In summary, fairness, privacy and security are generally recognized as core technical attributes of AI systems. Explainability and openness can be collectively referred to as transparency. Some researchers have further expanded the scope of relevant principles and highlighted the importance of human autonomy, oversight, environmental well-being and sustainability, indicating that the conceptual framework is constantly evolving and pointing out new directions for future research in this field. Forward-looking tech ethics guidelines encourage stakeholders to conduct ethical reviews throughout technological innovation. By comprehensively analyzing data governance, model development and decision support, we map and reframe the general principles of Responsible AI to the specific requirements of groundwater prediction, explore fair and responsible development and application of AI models across scenarios, and ultimately establish Responsible AI principles tailored for groundwater prediction models.

## 2.2 LSTM

Long Short-Term Memory (LSTM) is considered a variant of Recurrent Neural Networks (RNN)[11]. It adds three gating structures (forget gate, input gate, and output gate) to the RNN architecture. These gates act as information filters, allowing information to be stored in cells as memory units for a long time and learning the long-term dependencies between network inputs and outputs. The forget gate controls the retention of long-term information, the input gate controls the entry of new information, and the output gate determines the information to be output. Through these gates, LSTM can adaptively control the extent to which the model's output depends on input data at different times.

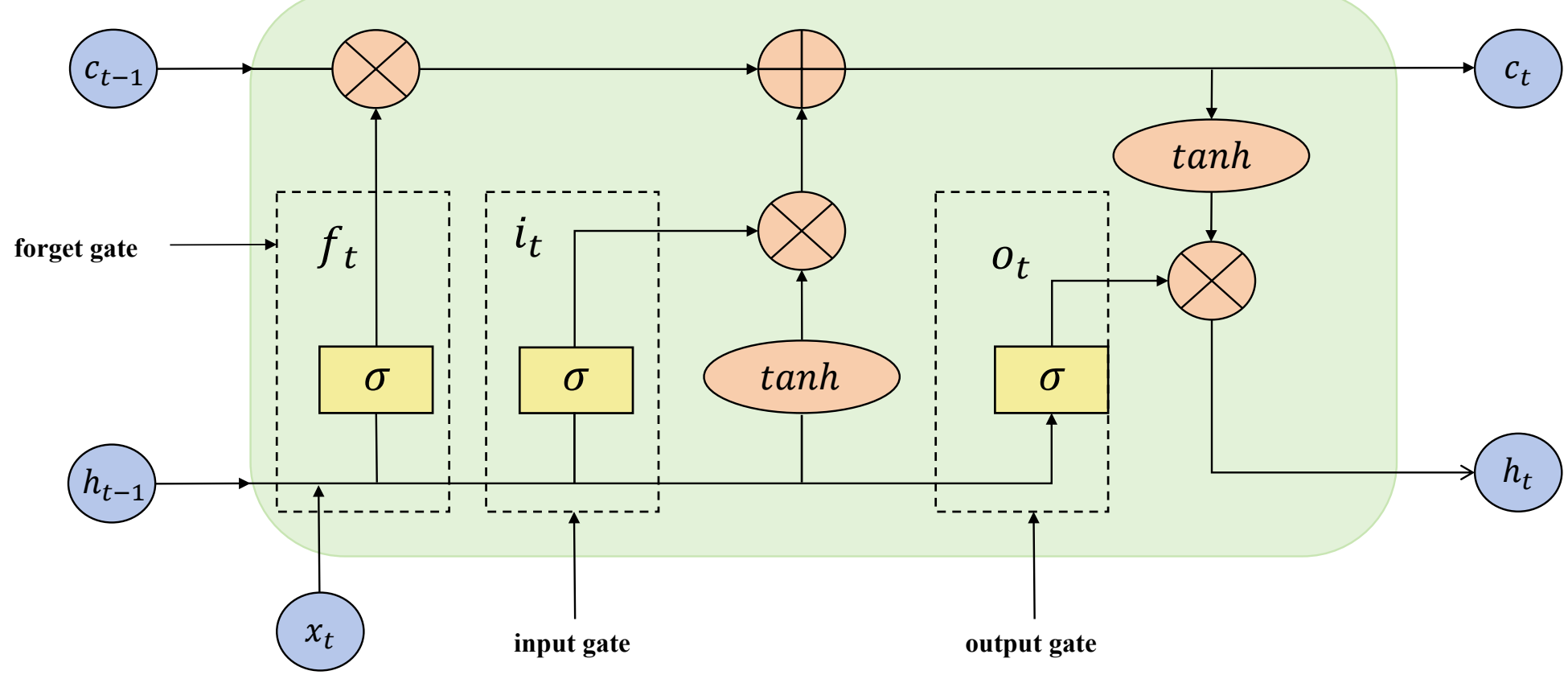

**Fig. 1 Basic structure diagram of LSTM**

Fig. 1 illustrates the basic structure of the LSTM network, where $i_t$, $f_t$ and $o_t$ represent the input gate, forget gate and output gate are denoted respectively. The calculation process at each time step is as follows:

$$i_t = sigmoid\left(W_i \cdot \left[h_{t-1}\ ,\ x_t\right] + b_i\right) \tag{1.1}$$

$$f_t = sigmoid\left(W_f \cdot \left[h_{t-1}\ ,\ x_t\right] + b_f\right) \tag{1.2}$$

$$\tilde{c}_t = tanh\left(W_c \cdot \left[h_{t-1}\ ,\ x_t\right] + b_c\right) \tag{1.3}$$

$$c_t = f_t \cdot c_{t-1} + i_t \cdot \tilde{c}_t \tag{1.4}$$

$$o_t = sigmoid\left(W_o \cdot \left[h_{t-1}\ ,\ x_t\right] + b_o\right) \tag{1.5}$$

$$h_t = o_t \cdot tanh\left(c_t\right) \tag{1.6}$$

Here, $W_i$, $W_f$, $W_c$, and $W_o$ denote weight parameters, while $b_i$, $b_f$, $b_c$, and $b_o$ denote bias parameters. $x_t$, $h_t$, $\tilde{c}_t$, and $c_t$ represent the input, hidden state, candidate cell state, and cell state at time step $t$ respectively.

## 2.3 Transformer

The main structure of the Transformer network consists of an encoder and a decoder (Fig. 2)[12]. The encoder consists of multiple encoder layers, each encoder layer contains two sub-modules: a multi-head attention layer and a fully connected feed-forward network layer. As the first sublayer in the encoding layers, the multi-head self-attention mechanism enables the model to interact with and focus on different positions in the input sequence during the encoding process. The mechanism uses three separate linear transformations to project the input sequence into three vector spaces: Q (query), K (key), and V (value). Each input vector calculates its similarity scores with other positions in the sequence to determine the importance of each position to other positions. Then, the weighted sum of all position vectors is obtained based on these scores, resulting in representations of each position relative to others. By performing parallel computations and concatenation across multiple attention heads, increases the expressive power and generalization ability of the model. The calculation process is:

$$MultiHead(Q,K,V) = Concat(head_1, head_2, \cdots, head_h)W^O \tag{1.7}$$

$$head_i = Attention(QW_i^Q, KW_i^K, VW_i^V) \tag{1.8}$$

Where $W^Q$, $W^K$, and $W^V$ are learnable weight matrices, $h$ is the number of attention heads. $W_i^Q$, $W_i^K$, and $W_i^V$ are the learnable parameter matrices of the separate attention heads.

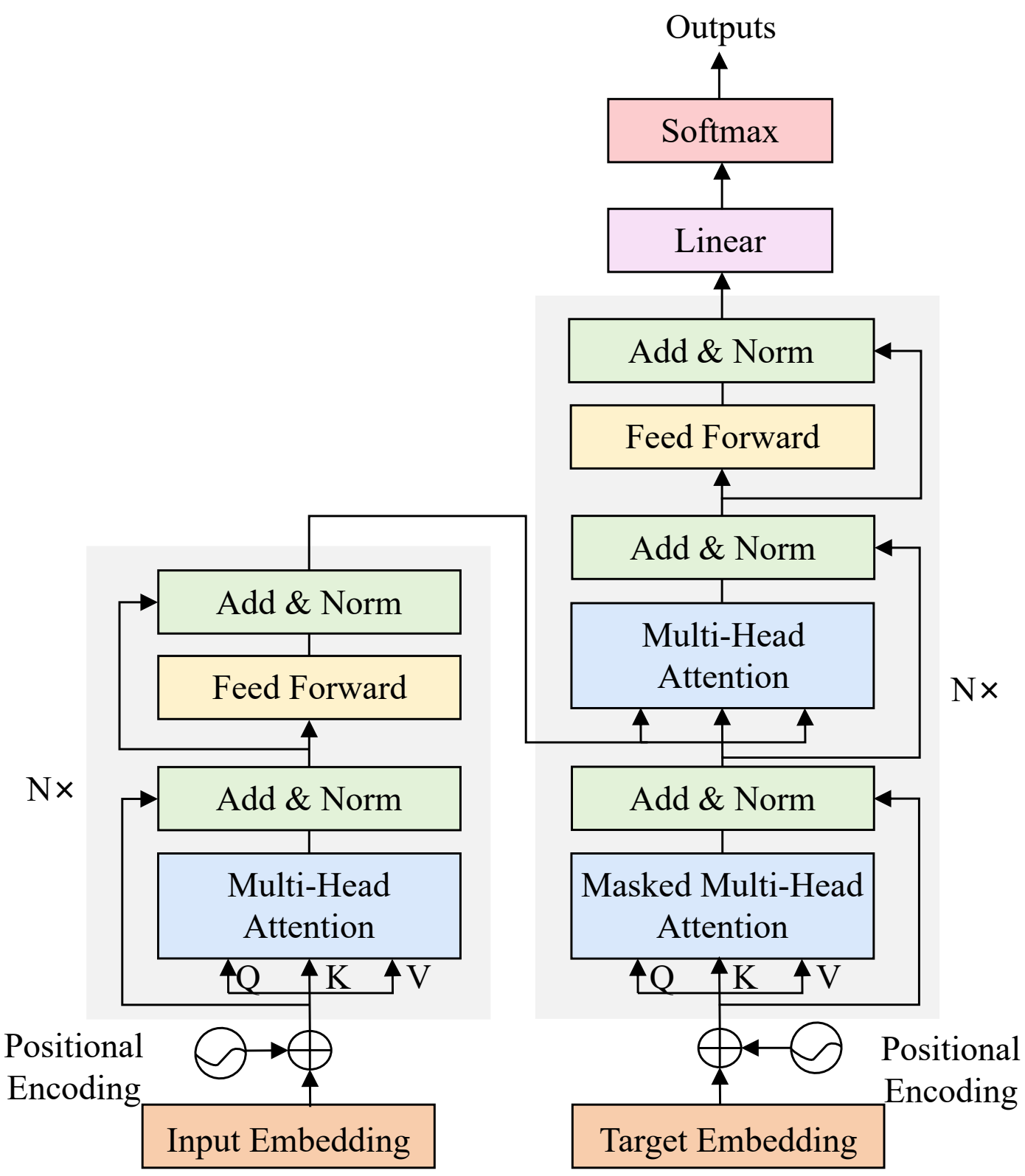

**Fig. 2 Diagram of Transformer structure**

The position encoding in the Transformer model enables encoding operations for the positions of each time step in the input sequence, modeling the positional information within the sequence. The position encoding is added to the input embedding representation, where for each position in the input sequence, the position encoding vector is composed of a combination of sine and cosine functions, with each dimension corresponding to a different period. The model is calculated as follows for the positional encoding:

$$PE(pos,2i)=sin\left(\frac{pos}{1000^{2i/d_{model}}}\right) \tag{1.9}$$

$$PE(pos,2i+1)=cos\left(\frac{pos}{1000^{2i/d_{model}}}\right) \tag{1.10}$$

Where $PE$ stands for the matrix position encoding matrix obtained from the position encoding operation, $i$ stands for dimension index, $pos$ stands for position index, and $d_{model}$ stands for dimension of input.

The feed forward neural network is placed between multi-head self-attention modules, adding nonlinear transformations to the model and enhancing the representation ability in high-dimensional feature space. The feed forward network consists of two fully connected layers, with the first layer using a ReLU activation function and the second layer without any activation. This network performs forward computation independently on the input at each position, allowing the model to capture local features of the input sequence more effectively.

## 2.4 Evaluation Indicators

Three metrics are adopted to evaluate the model prediction performance: Mean Absolute Error (MAE), Root Mean Square Error (RMSE), and Coefficient of Determination ($R^2$). MAE is the average of the absolute errors between predicted and observed values, RMSE is the square root of the mean of the squared errors between predicted and observed values, and $R^2$ measures the proportion of data variance explained by the model, which is defined as:

$$MAE=\frac{1}{n}\sum_{i=1}^{n}\left|y_i-\hat{y}_i\right| \tag{1.11}$$

$$RMSE=\sqrt{\frac{1}{n}\sum_{i=1}^{n}\left(y_i-\hat{y}_i\right)^2} \tag{1.12}$$

$$R^2=1-\frac{\sum_{i=1}^{n}\left(y_i-\hat{y}_i\right)^2}{\sum_{i=1}^{n}\left(y_i-\overline{y}_i\right)^2} \tag{1.13}$$

Where $n$ is the sample size, $y_i$ and $\hat{y}_i$ are the observed and predicted values, $\overline{y}_i$ is the mean of the observed values. Smaller MAE and RMSE values indicate lower prediction errors and higher model accuracy. A larger $R^2$ value reflects better model fit to the data: when R2=1, the model perfectly explains all the variance in the data; a negative $R^2$, however, indicates extremely poor prediction performance.

# 3 Responsible AI Framework for Groundwater Modeling

Throughout the entire lifecycle of using AI models to predict groundwater levels, multiple groups are affected directly or indirectly. To ensure the operability and social adaptability of the framework, this paper identifies four key stakeholders in groundwater prediction: data providers, model developers, model users, and local communities and the general public. As shown in Table 1, each party holds distinct responsibilities regarding data resources, model performance, water resource utilization and ecological protection. These responsibilities define their overall roles and functions within the Responsible AI framework. Meanwhile, the Responsible AI practices elaborated in Section 3.2.5 refine task division and clarify the entities responsible for specific implementation. The two aspects complement each other to jointly advance the scientificity, ethics and sustainability of Responsible AI applied to groundwater management.

**Table 1 Duties of all stakeholders**

| Stakeholders | | Duty |
|---|---|---|
| Data Providers | | Provide multi-source data, such as hydrology, geology and meteorology; |
| Model Developers | | Design, develop, train, validate, and deploy AI models for groundwater;<br>Introduce explainability and bias detection mechanisms;<br>Maintain model performance and security. |
| Model Users | Water Affairs Bureau | Lead water resource allocation, emergency dispatch, and interregional coordination;<br>Supervise drinking water safety. |
| | Ministry of Agriculture | Establish and adjust agricultural irrigation quotas;<br>Optimize the efficiency of agricultural water use. |
| | Environmental Protection Industry | Monitor ecological flows and track pollution incidents;<br>Promote environmental justice and public participation;<br>Promote green policymaking. |
| Communities & Public | | Daily water safety;<br>Ecological protection;<br>Taking part in public deliberation. |

## 3.1 Transparency

Transparency means that the internal logic of a model can be presented in a human-understandable manner, with full-process recording, reproducibility and verifiability, so as to improve the credibility and acceptability of decisions[13].

### 3.1.1 Explainability

Explainable Artificial Intelligence (XAI) is a core enabler for the practical implementation of Trustworthy AI, as well as an essential means to interpret the internal logic of black-box models. Black-box models fail to predict their performance under extreme or unseen scenarios, and cannot quickly identify and fix errors once they occur. By identifying which input factors influence the decisions of complex black-box algorithms, we can gain a holistic view of model mechanisms. Combined with traceability systems and clear interpretation tailored for target audiences, XAI can substantially improve model trustworthiness.

Since interpretation effectiveness largely depends on the audience and application goals, researchers have developed a variety of XAI techniques for classification, regression and other tasks[14-16]. In terms of model dependency, they fall into model-agnostic and model-specific approaches: the former applies to all machine learning models, while the latter is designed for specific model architectures. By interpretation timing, they are divided into intrinsic explainability and post-hoc explainability. The former builds transparent models via interpretable design, whereas the latter uses auxiliary tools to interpret model behavior after training. From the perspective of interpretation scope, there are global and local explainability methods. Global methods focus on the overall behavioral patterns of a model, while local ones analyze the causes of individual prediction results. Groundwater level variations are jointly affected by hydrological conditions, meteorological factors and human activities, whose impacts also vary significantly across time series. Enhancing model explainability ensures that the prediction process is interpretable, reviewable and clearly communicable to managers and decision-makers. It not only aligns model predictions with actual hydrological mechanisms, but also makes the prediction results consistent with regional hydrological rules.

### 3.1.2 Traceability

Traceability consists of mechanisms and procedures for tracking system data, development and deployment via documented identifiers. In practice, metadata management, structured logging, version control and distributed storage are adopted to record data sources, processing flows, model parameters and decision logic throughout the lifecycle, guaranteeing the whole process reviewable, verifiable and reproducible. Embedding traceability and logging in early AI design facilitates model auditing and satisfies diverse transparency demands.

Groundwater predictions greatly influence water resource management and scheduling. A complete traceability system enhances model reliability and transparency, and supports accountability and error tracing. This paper therefore proposes a full-process recording mechanism. Log systems document operations across all stages: data collection and preprocessing details, model structure, variable selection, dataset partitioning, hyperparameters and training time, as well as prediction results and evaluation metrics. Since centralized logs are vulnerable to tampering, blockchain is recommended for core data storage. All parameter adjustments during model debugging are recorded on immutable distributed ledgers, effectively preventing falsification and boosting trust in data and model decisions.

## 3.2 Privacy and Data Governance

Privacy is about respecting and protecting the rights of individuals, making sure personal data is kept safe and private, not misused or accessed without permission, and giving people control and knowledge over how their data is used. Data governance, on the other hand, focuses on data quality, making sure data stays complete and confidential, and is used legally, transparently, and fairly throughout its lifecycle. Together, they set management standards for data access, quality, and privacy[17].

Multiple technical approaches can achieve effective data protection. Federated learning retains raw data locally on each client, performs early edge aggregation, and exchanges only task-essential information. Meanwhile, stochastic gradient descent-based differential privacy protects privacy in deep neural network training via gradient clipping and Gaussian noise injection. Applied individually or jointly, these methods safeguard personal privacy and minimize potential risks. Data governance further establishes a standardized data usage framework covering four key aspects: sensitive data governance adheres to data minimization, de-identification and hierarchical protection strategies to prevent leakage and misuse; fine-grained access management allocates role-based permissions and adopts end-to-end encryption to resist unauthorized access and tampering; strict data quality control ensures data accuracy, completeness and timeliness prior to model training; comprehensive regulatory compliance clarifies data usage boundaries to guarantee legal, fair and trustworthy data application.

## 3.3 Technical Robustness and Safety

Faced with multiple challenges including multi-source heterogeneous data,

monitoring errors, complex aquifer structures, climate fluctuations and potential malicious attacks, the model must maintain stable and reliable outputs. While ensuring robustness, a responsible AI system also needs to guarantee security, namely the capability to guard against potential risks during operation.

### 3.3.1 Technical robustness

Robustness focuses on prevention rather than post-event remediation. Proactive risk prevention strategies, combined with uncertainty quantification and out-of-distribution detection, can enhance system security. It correlates robustness with generalization performance under adversarial and unseen scenarios, emphasizing stable performance even under worst-case conditions and out-of-distribution inputs, rather than merely average accuracy. Groundwater time series data commonly suffer from sensor errors, noise interference and missing observations, which may structurally impair model predictions. Therefore, potential risks arising from data collection, transmission and preprocessing should be identified and addressed prior to model training to ensure training on high-quality datasets. Additionally, extreme climate events or intense human activities such as sharp rises in rainfall or water extraction may lead to unreliable outputs. It is necessary to quantify model uncertainty in advance and inform decision-makers of prediction confidence. Only models with strong robustness can deliver consistent and reliable results in complex real-world environments and effectively avoid prediction errors.

### 3.3.2 Safety

The required protection level depends on model risks, determined by prediction performance and potential losses from wrong decisions. For instance, underestimated groundwater decline in over-exploited irrigation areas may cause ecological degradation, while prediction errors in water supply zones will threaten domestic water safety. Thus, accuracy shall be the core security objective, guaranteed via continuous validation, testing and monitoring. Meanwhile, encryption can protect sensitive hydrological data against leakage and attacks, sustaining stable system operation.

### 3.3.3 Reproducibility

Reproducibility crisis has emerged as a major bottleneck restricting scientific progress in AI, referring to whether an AI experiment exhibits the same behavior when repeated under the same conditions. To avoid confusion between the terms, separate definitions are provided[18]:

- *Repeatability:* the ability of the research team to obtain consistent results using the same experimental setup.

- *Reproducibility:* an independent team validates original findings by following documented experimental settings. It includes dependency reproducibility (using original code and data) and independent reproducibility (re-implementing experiments based on the published description), with the latter being more valued for its rigor.
- *Direct Replicability:* an independent team intends to change the way the experiment is implemented, while keeping the hypothesis and experimental design consistent with the original study, in order to test the stability of the results.
- *Conceptual Replicability:* an independent team examines the same hypothesis through entirely new experimental approaches.

The extent to which these evaluations can be achieved depends on the confidentiality level of the system and the proprietary nature of the development platform, as well as other constraints. In open research, publicly released code enables third-party replication. In commercial software, where core algorithms or datasets are undisclosed, reproducibility relies on strict protocols, detailed documentation, and standardized testing, ensuring consistent evaluation results and credible assurance of system robustness and safety.

## 3.4 Fairness

A fair AI model should reduce biases that might come up in data, algorithm design, or application scenarios, and avoid discriminating against specific individuals or groups[19].

### 3.4.1 Data bias

Data bias refers to systematic errors in datasets, stemming from unrepresentative training data and inherited historical inequities. For example, historically low female representation in recruitment data may lead algorithms to establish false links between gender and job suitability, resulting in systematic discrimination against female applicants during resume screening. In groundwater prediction, water-scarce regions tend to have more detailed and frequent observations, while remote or water-abundant areas often suffer from sparse and incomplete records due to cost and technical constraints. Such data imbalance makes models biased toward well-sampled regions and weakens prediction performance in data-deficient areas. Besides, input variables need screening and optimization for different climatic zones: precipitation exerts limited impact in arid areas, where other water-level influencing factors take priority, whereas it acts as a core variable in humid regions.

Data fairness can be achieved through proper data collection and feature selection.

Statistical methods including regression analysis and T-test, together with data augmentation techniques such as SMOTE, can be applied to mitigate data bias.

### 3.4.2 Algorithmic bias

Algorithm bias arises from model architecture, algorithm design or optimization strategies, rather than data or external factors, and may also occur during algorithm design and user interaction. Models vary greatly in adaptability to data distribution, time series length and nonlinear features. For example, LSTM and GRU excel at capturing local temporal dependencies and nonlinear patterns, while Transformer relies on large datasets and long sequences for global attention learning, and tends to suffer from overfitting and poor generalization with limited samples. Popularity-based recommendation algorithms prioritize high-popularity content and marginalize the needs of minority groups. Improper evaluation metrics may also drive algorithms to improve overall performance at the expense of minority interests. During model training, fairness constraints or penalty terms can be incorporated into optimization objectives. Bayesian optimization, grid search and particle swarm optimization can globally tune hyperparameters such as regularization coefficients, learning rates and iterations. Traversing preset hyperparameter combinations avoids subjective bias from manual tuning. Moreover, feature complexity affects model structure. Reducing model scale can mitigate overfitting and maintain prediction accuracy. Selecting proper architectures based on data characteristics and adopting fair training strategies can effectively reduce algorithm bias and ensure balanced and fair groundwater prediction across all regions.

## 3.5 Accountability

Accountability clarifies the responsibilities of all parties, ensuring traceable decisions, auditable conduct and accountable outcomes. In human-led decisions, people bear corresponding liabilities; in environment-restricted systems, regulatory bodies take charge; while stochastic decision mechanisms lack clear accountability, as they cannot explain decision logic or assign liabilities to specific entities. Essentially, accountability serves as a moral constraint that guides practitioners to uphold fairness, environmental justice and prudence during model design and deployment. It also acts as an institutional mechanism with standardized tools and workflows to implement liability rules. Closely linked to fairness and security, it defines who shall be held responsible for technical risks and biased outputs, even when such issues can be fixed technically.

To embed accountability into groundwater AI modeling, responsibilities among stakeholders must be explicitly delineated to align technical practices with ethical objectives. Building on the four primary stakeholder groups identified earlier, we propose a RACI matrix (Table 2) to formalize responsibility allocation across five critical stages: data quality, model development, water resource scheduling, ecological impact assessment and grievance redress. This will establish a transparent chain of responsibility and prevent it from becoming diluted. As shown in the table below, in the data quality phase, data providers are responsible (R) and accountable (A) for results, while model developers offer technical support (C). Model users provide advice (C), and the public is informed (I).

- *Responsible:* entities directly implementing tasks (e.g., data providers ensuring accuracy, model developers optimizing performance).
- *Accountable:* stakeholders ultimately responsible for outcomes (e.g., water resources bureau for allocation policies, model developers for predictive accuracy).
- *Consulted:* actors offering expertise or feedback (e.g., environmental organizations, communities providing water usage insights).
- *Informed:* stakeholders notified of decisions but not involved in implementation (e.g., public receiving reports on predictions or water security).

**Table 2 RACI Matrix for Responsibility Determination in Groundwater AI Modeling**

**(R = Responsible，A = Accountable，C = Consulted，I = Informed)**

| Task | Data Providers | Model Developers | Model Users | | | Communities & Public |
|---|---|---|---|---|---|---|
| | | | Water Conservancy Bureau | Ministry of Agriculture | Environmental Protection Industry | |
| Data Quality | R/A | C | C | C | C | I |
| Model Development | I | R/A | C | C | I | I |
| Water Allocation & Emergency Response | I | C | R/A | C | C | I |
| Agricultural Irrigation Quotas | I | I | C | R/A | C | I |
| Ecological Impact Evaluation | I | C | R/A | C | R | C |
| Complaint Handling & Remediation | C | C | A | C | C | R |

## 3.6 Sustainability

The wide application of AI models should strictly follow the principles of green and low-carbon development to minimize energy consumption and carbon emissions. Meanwhile, AI-powered models and decision support systems shall reduce disturbances to natural ecosystems, prevent excessive groundwater extraction and ecological imbalance, and maintain the stability of aquifer systems and sustainable ecological water use.

### 3.6.1 Computational sustainability

Computational sustainability means prioritizing methods with low energy consumption and high accuracy during model development and deployment. Model structure optimization, training strategy adjustment and rational allocation of computing resources are adopted to cut energy use and computational overhead in training and inference. For example, adopting parameter-efficient network architectures combined with transfer learning avoids resource waste from repeated training from scratch. Model pruning, knowledge distillation and early stopping strategies reduce inference computation and improve efficiency without obvious accuracy loss.

This concept also applies to groundwater time series prediction. Models balancing accuracy and computational efficiency are built, and green indicators such as energy consumption and computing costs are incorporated into the evaluation system alongside traditional metrics like RMSE and $R^2$, so as to strike a balance between model performance and sustainability. Achieving sustainable AI requires full-stack optimization across data, algorithms, training and hardware. It aims to balance accuracy and environmental costs to determine the optimal model architecture and parameter settings.

### 3.6.2 Humans and ecological sustainability

Groundwater prediction pursues rational water utilization while balancing social and ecological benefits. AI models should coordinate multi-stakeholder water demands to ensure equitable water allocation and avoid adverse impacts on vulnerable groups and ecosystems. In water-scarce regions, models prioritize groundwater level maintenance, efficient irrigation and ecological restoration to sustain the dynamic balance between groundwater recharge and extraction. By embedding ecological water demand thresholds and scenario simulation, the models can predict long-term groundwater variations and extreme water level fluctuations, supporting groundwater regulation and interregional water allocation to prevent ecological degradation. Dynamically optimized models adapt to varying watershed conditions and promote ecological improvement. This complies with Responsible AI principles, which prioritize public welfare and ecological security rather than merely pursuing model performance at the expense of ecological and social interests.

# 4 Experiment

## 4.1 Study sites and data

The middle reaches of the Heihe River Basin (97°20′–102°12′ E, 37°28′–39°57′ N; Fig. 3) lie in the central Hexi Corridor. With well-developed artificial irrigation systems, this area is a major water consumption zone. It features a typical temperate continental arid climate, with an annual average temperature of 7–9 °C and annual precipitation of 100–200 mm, characterized by arid conditions, scarce rainfall and intense evaporation. The terrain slopes downward from east to west and from south to north, covering Zhangye City, Linze County and Gaotai County in Gansu Province. Dominated by alluvial fans, the landform presents a distinct zonal distribution from south to north. The southern Qilian Mountains form a piedmont alluvial-proluvial gravel plain with high stratum permeability, acting as the primary recharge zone where surface water infiltrates into groundwater. The northern part transitions into an alluvial-proluvial fine-soil plain in the central basin. Boasting fertile soil suitable for farming, this area is home to concentrated artificial oases and irrigated farmland.

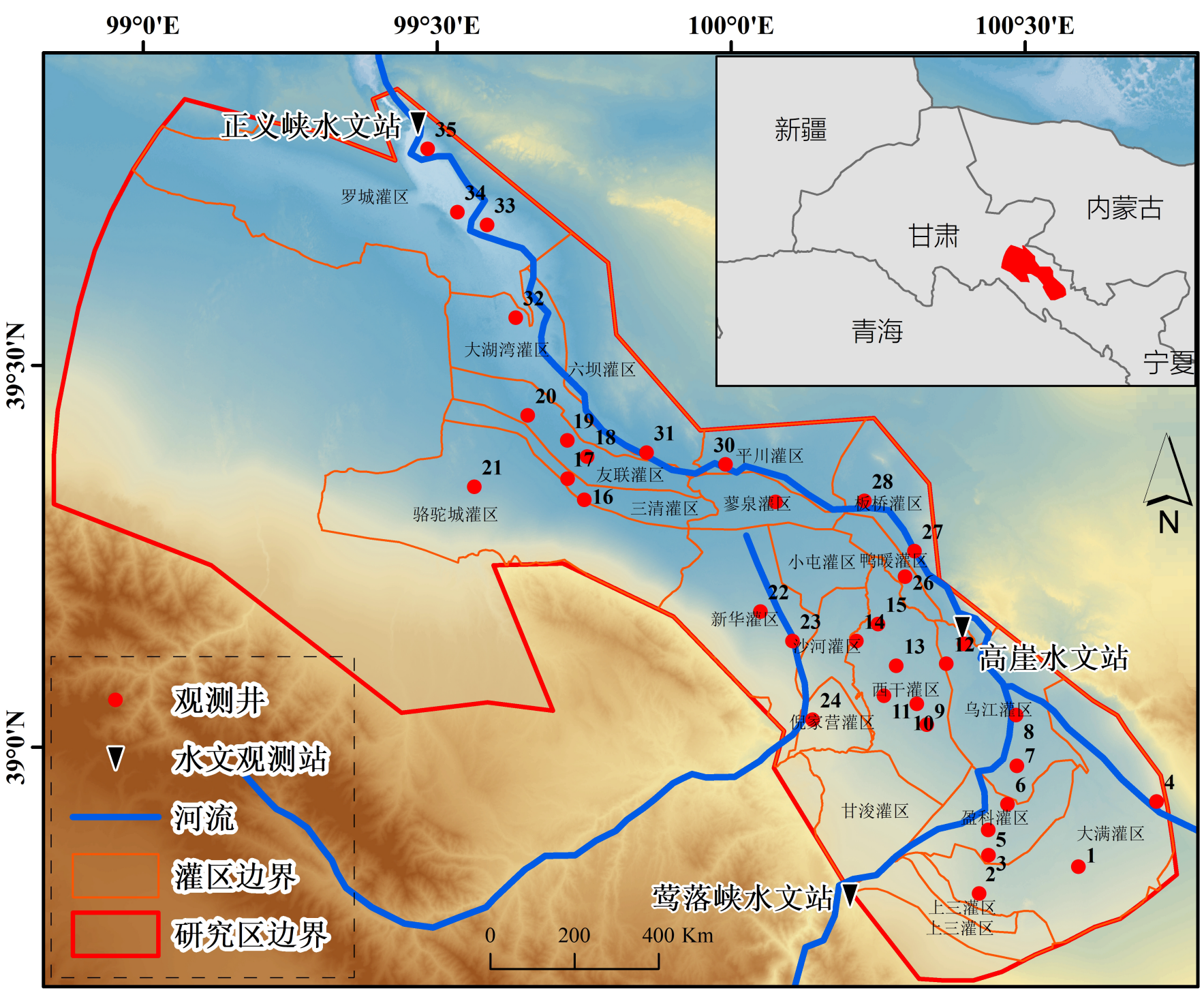

**Fig. 3 Geographical location and topographical features of the study area.** (Detailed topographical map of the study area, illustrating the administrative boundaries, main river networks, and the spatial distribution of hydrologic stations and groundwater observation wells)

This study adopts multi-source data of the middle reaches of the Heihe River Basin from 1986 to 2008 (Table 3). The hydrological datasets include groundwater level, recharge depth and water extraction data. Meteorological data such as precipitation and temperature are obtained from the National Meteorological Science Data Center (https://data.cma.cn/). All the above data are on a monthly scale.

**Table 3 Data used in the experiment**

| Data name | Spatial resolution | Temporal resolution | Unit | Period |
|---|---|---|---|---|
| Pumping | Irrigation | Monthly average | $m^3$ | 1986-2008 |
| Recharge depth | | | m | |
| Temperature | Station | | 0.1℃ | |
| Precipitation | | | 0.1mm | |
| Groundwater level | | | m | |

## 4.2 Data processing

Due to the unique observation conditions and long collection cycles of groundwater monitoring data, the datasets vary greatly in magnitude and dimension. Direct training of neural network models with raw data may lead to overfitting or failure to capture effective features. Therefore, this paper conducts systematic data quality inspection prior to model development, which serves as the prerequisite for ensuring model robustness and the core of data governance.

Linear interpolation is applied to fill missing values and maintain the continuity and overall trends of time series. The box plot method based on the interquartile range (IQR) is used to remove outliers that deviate markedly from statistical distribution and actual hydrological processes. To eliminate the impact of dimensional differences on training stability, Max-Min normalization is adopted to standardize qualified data. All variables are scaled to a unified range while retaining their original variation trends, and the predicted results are inversely normalized to the original physical dimensions for comparison with measured data. The original monthly dataset contains 276 samples, which are expanded to 841 samples via spline interpolation to facilitate neural network modeling and improve model interpretability. The full dataset is chronologically divided into a 70% training set, a 10% validation set and a 20% test set. A sliding time window with a window size of 60 is adopted for sample reconstruction to realize one-step prediction of groundwater levels for the next day.

## 4.3 Model Configuration and Training

The LSTM network consists of one LSTM layer, one Dropout layer and one fully connected layer. The Transformer adopts a two-layer encoder-only architecture, and each layer is composed of a self-attention layer, a feed-forward neural network layer, a Dropout layer, as well as a residual connection and normalization layer.

The LSTM model uses a grid search method, while the Transformer model uses Bayesian optimization, systematically exploring and evaluating their key hyperparameters to select stable configurations, thus creating a structured and traceable mechanism for fair hyperparameter selection. The maximum number of iterations is set to 150. Both models adopt 2-fold cross-validation and an early stopping strategy with a patience value of 20. The validation set error is monitored in real time, and training will stop automatically if no performance improvement is observed over consecutive iterations. This strategy works in tandem with the Dropout layers embedded in the network, it prevents the model from overfitting to noise by limiting training duration and reduces reliance on specific neuron combinations through network design. The optimal hyperparameter combinations of the two models at typical observation wells are listed in Table 4 and Table 5.

**Table 4 Range of hyperparameters and optimal combination for LSTM model**

| Hyperparameter | Scope | Observation Well | |
|---|---|---|---|
| | | 2 | 28 |
| Learning Rate | [0.001，0.01，0.1] | 0.01 | 0.01 |
| Batch Size | [32，64] | 32 | 32 |
| Optimizer | [Adam，RMSprop] | Adam | Adam |
| Neuron Number | [16，32，64] | 32 | 64 |
| Dropout Rate | [0.1，0.2，0.3] | 0.1 | 0.2 |

**Table 5 Range of hyperparameters and optimal combination for Transformer model**

| Hyperparameter | Scope | Observation Well | |
|---|---|---|---|
| | | 2 | 28 |
| Model Feature Dimension | [16，128][16] | 16 | 128 |
| Attention Heads | [1，6][1] | 3 | 1 |
| FFN Dimension | [16，64][16] | 32 | 16 |
| Learning Rate | [$10^{-4}$，$10^{-1}$][log] | 0.01 | 0.0001 |
| Batch Size | [16，64][16] | 32 | 16 |
| Dropout Rate | [0.1，0.3][0.1] | 0.2 | 0.2 |

Figure 4 presents the groundwater level prediction results for typical observation wells. Affected by regional differences in hydrological conditions, the two models show distinct prediction performance. Single-variable models fail to capture the nonlinear variations of groundwater levels relying merely on historical water level data, resulting in obvious prediction errors. After integrating multi-source features, the prediction accuracy of both models for the two wells is improved. In terms of model adaptability, LSTM performs steadily in both regions and is good at capturing local abrupt changes in groundwater time series. Benefiting from the global self-attention mechanism, Transformer achieves overall better prediction performance after fully learning sequential dependencies. It demonstrates stronger fitting and tracking capabilities, especially at the peaks and troughs of groundwater level fluctuations.

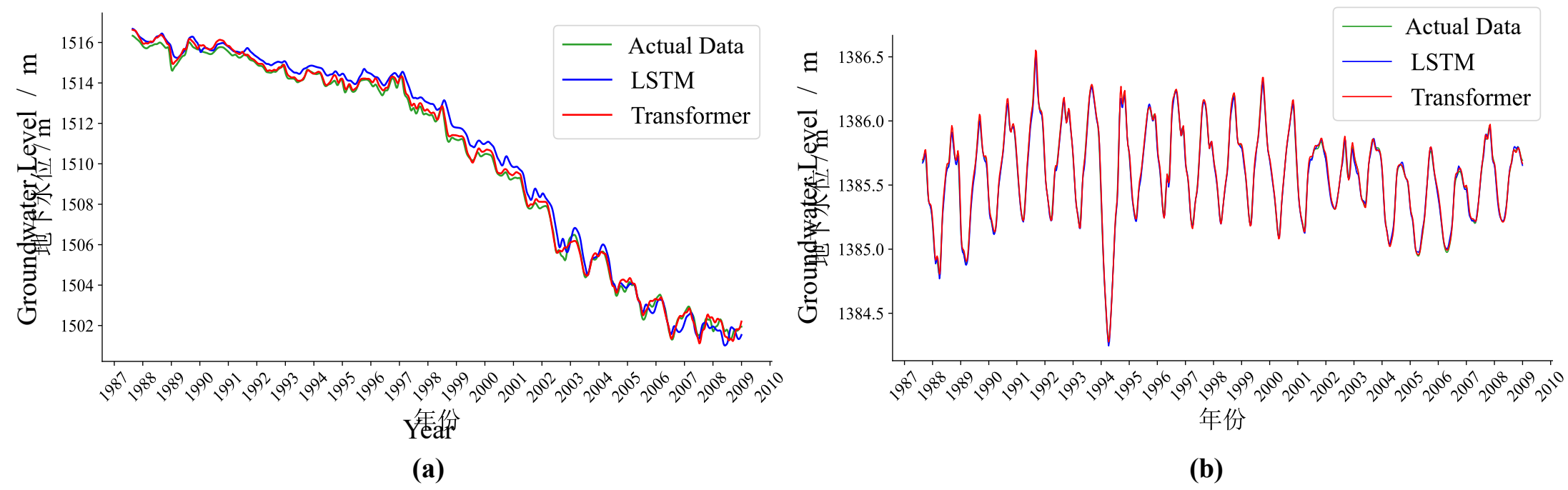

**Fig. 4 Comparison of actual and simulated groundwater levels for LSTM and Transformer models for (a) Observation well No.2; (b) Observation well No.28.**

The evaluation results of the two models are shown in Table 6. At Observation Well No.2, the Transformer model significantly outperforms the LSTM model, with an $R^2$ of 0.924, approximately 11% higher than that of LSTM, indicating that Transformer has higher fitting accuracy and stronger explanatory power for water level variations. At Observation Well No.28, the performance difference between the two models is small: the error metrics of LSTM are slightly lower, and its $R^2$ (0.989) is also marginally higher than Transformer's 0.986, with both achieving high prediction accuracy. These results demonstrate that the global attention mechanism of the Transformer model offers greater advantages at wells with complex hydrological conditions and large water level fluctuations, making it better suited to meet the dual requirements of accuracy and stability set by Responsible AI for high-risk hydrological prediction scenarios.

**Table 6 Results of evaluation metrics of different models on the test set**

| Observation Well | Evaluation Metrics | LSTM | Transformer |
|---|---|---|---|
| 2 | MAE | 0.280 | 0.174 |
| | RMSE | 0.342 | 0.202 |
| | $R^2$ | 0.815 | 0.924 |
| 28 | MAE | 0.018 | 0.034 |
| | RMSE | 0.021 | 0.038 |
| | $R^2$ | 0.989 | 0.986 |

# 5 Results and discussions

## 5.1 Explainability Analysis

The SHAP method is adopted to conduct interpretability analysis on the prediction results of the LSTM and Transformer models by randomly selecting 200 representative samples. The marginal contribution of each input feature to the model output is characterized by calculating Shapley values, thereby enabling post-hoc interpretation of complex deep learning models. The expression is given by:

$$g\left(z^{'}\right)=\phi_0+\sum_{j=1}^{M}\phi_j z_j^{'} \tag{1.14}$$

Where $g$ denotes the LSTM model, $z^{'}$ is the feature combination vector, $M$ is the number of input features, and $\phi$ is the calculated Shapley value.

From the SHAP results in Figure 4.9, both LSTM and Transformer identify the key physical drivers of groundwater change in the study area, with pumping volume and recharge depth as the main external factors, reflecting the dominant roles of human extraction and natural recharge. Precipitation and temperature show lower importance, consistent with the arid/semi-arid climate where meteorological factors have weak direct impacts. At Well 2, Transformer exhibits clearer and more stable attribution of pumping’s negative and recharge’s positive effects, while LSTM shows occasional jumps in feature contributions, leading to weaker consistency.

Overall, both models correctly capture the core drivers and physical mechanisms, but differ in interpretability. Transformer, using self-attention, achieves more balanced use of full temporal information, with more stable, traceable feature contributions, better meeting the transparency requirement for high-risk water management. LSTM, limited by its gating structure and local dependencies, tends to produce more interpretive noise and is less effective at capturing long-term hydrological patterns, making its interpretability inferior to Transformer.

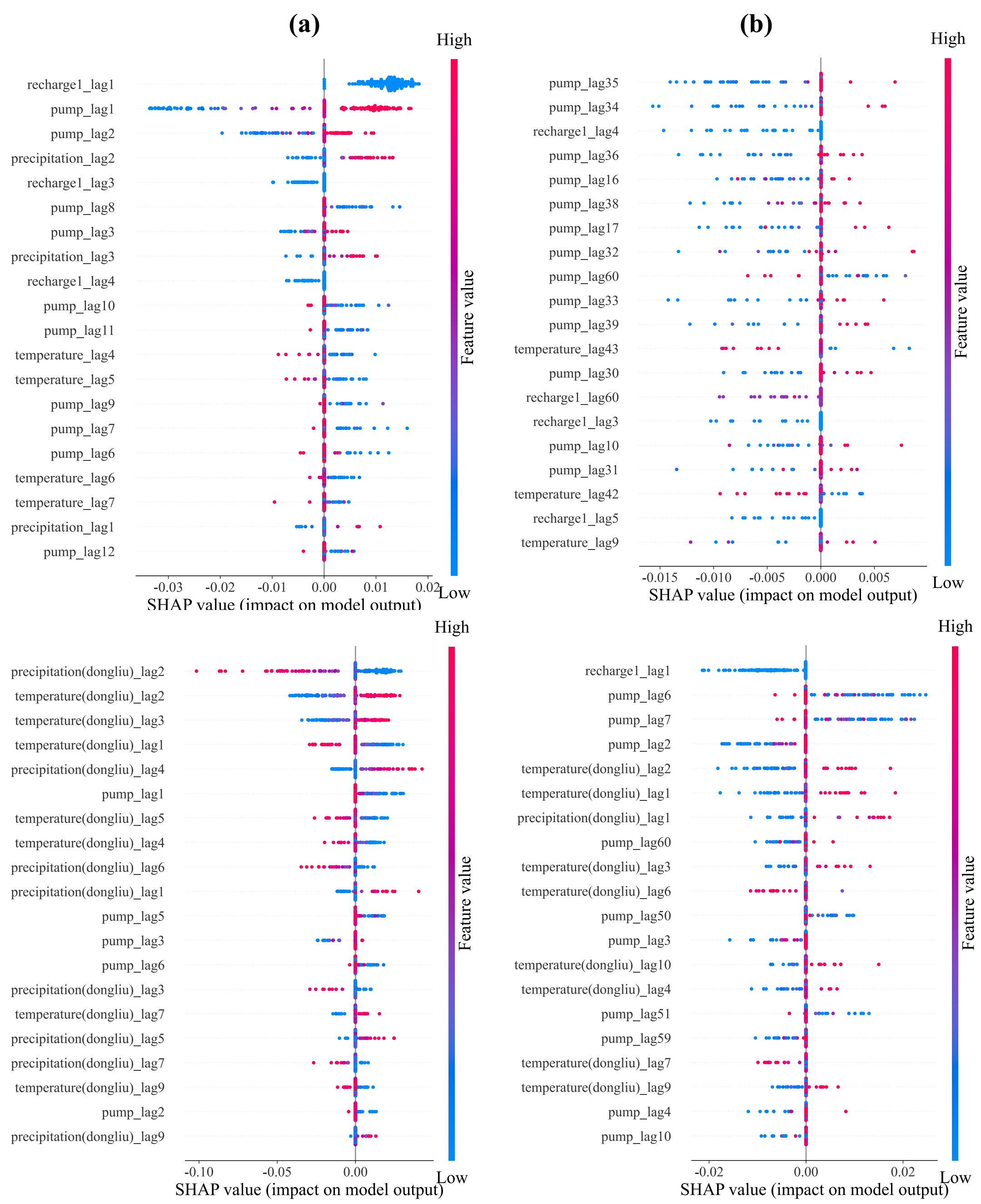

**Fig. 5 SHAP interpretable analysis results of observation wells No.2 (upper) and No.28 (lower) for (a) LSTM model; and (b) Transformer model.**

## 5.2 Technical Robustness Analysis

Based on the Monte Carlo simulation method, the uncertainty of model predictions is quantitatively analyzed. Normally distributed noise with a mean of 0 and a standard deviation of 0.05 is added to the input features to simulate monitoring errors. In each simulation run, the perturbed data are fed into the optimal model trained in Section 4.3 to generate predictions. To balance efficiency and stability, a total of 250 simulations are performed. Finally, the mean and standard deviation of predictions at each time point are calculated from the results, where the standard deviation represents the magnitude of uncertainty, and the confidence intervals provide quantifiable error bounds for decision-making.

The Monte Carlo simulation results of the LSTM and Transformer models are shown in Fig 6. For both wells, the mean predictions (red lines) of both models generally align with the observed groundwater levels (green lines), and the 95% confidence intervals (shaded areas) cover most measured values. However, the LSTM model exhibits a gradually widening confidence interval and obvious error accumulation over time at Well No.2, while the Transformer maintains a narrower interval and better fits the long-term declining trend. At Well No.28, both models capture the seasonal fluctuations, but the Transformer shows tighter confidence bounds, reflecting more accurate depiction of the water level variations.

The temporal changes in prediction standard deviation further confirm the above findings. The LSTM model yields average standard deviations of 0.2684 m for Well No.2 and 0.2304 m for Well No.28, with significant upward trends and unstable fluctuations, indicating high and growing uncertainty. In contrast, the Transformer achieves lower average standard deviations of 0.2094 m and 0.1277 m for the two wells, respectively, and effectively suppresses error accumulation. These results demonstrate that the Transformer model outperforms LSTM in controlling prediction uncertainty, especially under complex hydrological conditions, providing more reliable error bounds for risk-informed water resource management.

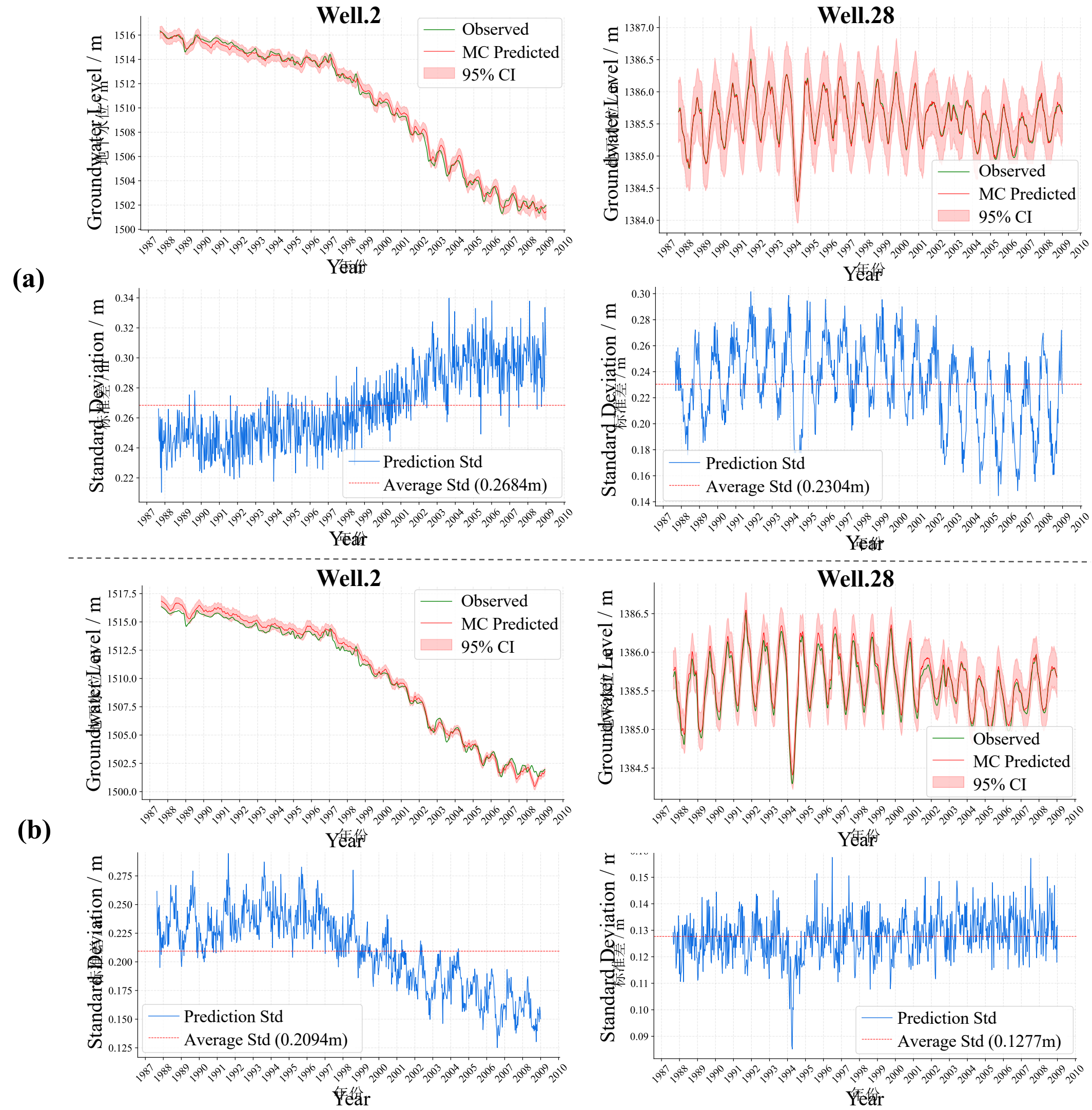

**Fig. 6 MC forecast mean plot (upper) and standard deviation time series plot (lower) for (a) LSTM model; and (b) Transformer model.**

## 5.3 Algorithmic Fairness Analysis

Taking Observation Well No. 2 as an example, we compare the model performance under three schemes (Table 8). Quantitative evaluation of LSTM and Transformer is conducted in Table 7 to verify the effectiveness of fair optimization and early stopping strategies.

**Table 7 Configuration solutions for three distinct scenarios**

| | Optimization Method | | Early stopping |
|---|---|---|---|
| | Grid Search | Bayesian Optimization | |
| Set One | × | × | × |
| Set Two | × | × | √ |
| Set Three | √ | √ | × |

**Table 8 Evaluation metrics of single variable models under different optimization strategies**

| | Model | Training Set | | | Test Set | | |
|---|---|---|---|---|---|---|---|
| | | MAE | RMSE | $R^2$ | MAE | RMSE | $R^2$ |
| Set One | LSTM | 0.317 | 0.376 | 0.948 | 2.133 | 2.386 | -7.045 |
| | Transformer | 0.993 | 1.074 | 0.823 | 2.412 | 2.524 | -9.093 |
| Set Two | LSTM | 0.893 | 0.947 | 0.862 | 0.859 | 0.877 | 0.172 |
| | Transformer | 0.647 | 0.746 | 0.915 | 1.154 | 1.30 | 0.167 |
| Set Three | LSTM | 0.673 | 0.692 | 0.927 | 0.681 | 0.693 | 0.339 |
| | Transformer | 0.589 | 0.628 | 0.939 | 0.380 | 0.439 | 0.595 |

Set one uses manually set parameters for LSTM and Transformer. While training set performance is decent, test set $R^2$ turns negative, indicating poor feature learning and large errors.

Set two retains the same parameters but adds early stopping to monitor validation errors. Though test set $R^2$ improves to over 15%, training error rises, and performance remains inferior to systematically optimized models.

Set three applies hyperparameter search (grid search for LSTM, Bayesian optimization for Transformer) without early stopping. Compared to Scheme 1, both models capture temporal features better, reducing errors and improving accuracy across the dataset.

In conclusion, directly setting model parameters has limitations and randomness, as the same parameters yield different prediction results on different datasets. Obtaining optimal fitting results would require manual trial of every combination, which is time-consuming. Automating this process with computer algorithms greatly reduces manual tuning time and ensures near-optimal hyperparameter combinations in a short period, forming the foundation for good model prediction performance. Meanwhile, the early stopping strategy can reduce computational cost, shorten training time, and effectively prevent overfitting to noisy data. Combining both strategies is essential to maximize model robustness and generalization ability.

## 5.4 Sustainability Analysis

The groundwater prediction models are built to ensure prediction accuracy and support environmental and ecological sustainability. Feature attribution shows pumping volume and recharge depth are key influencing factors. We adopt rolling prediction with trained LSTM and Transformer models: each predicted water level is used as the input for the next step to achieve continuous sequence forecasting. We analyze groundwater dynamics under different extraction and recharge conditions and set ecological risk thresholds using historical data to guide water resource management.

Using 2007–2008 data as the baseline, we establish one baseline scenario and two intervention scenarios to predict groundwater levels over the next 720 days (two years). The green, red and blue curves correspond to the baseline, a 10% gradual increase in pumping volume, and a 10% gradual increase in recharge depth respectively. In the baseline scenario, Well No.2 shows a continuous water level decline, while the water level of Well No.28 remains stable between 1384 m and 1387 m. Transformer produces more conservative and continuous predictions. Rising pumping intensity lowers peak water levels and accelerates the decline at Well No.2, proving the piedmont plain is highly sensitive to extraction changes. Increased recharge effectively lifts the water level of Well No.2 and relieves over-extraction pressure, yet exerts limited impact on Well No.28, where the river plain groundwater system maintains stable operation.

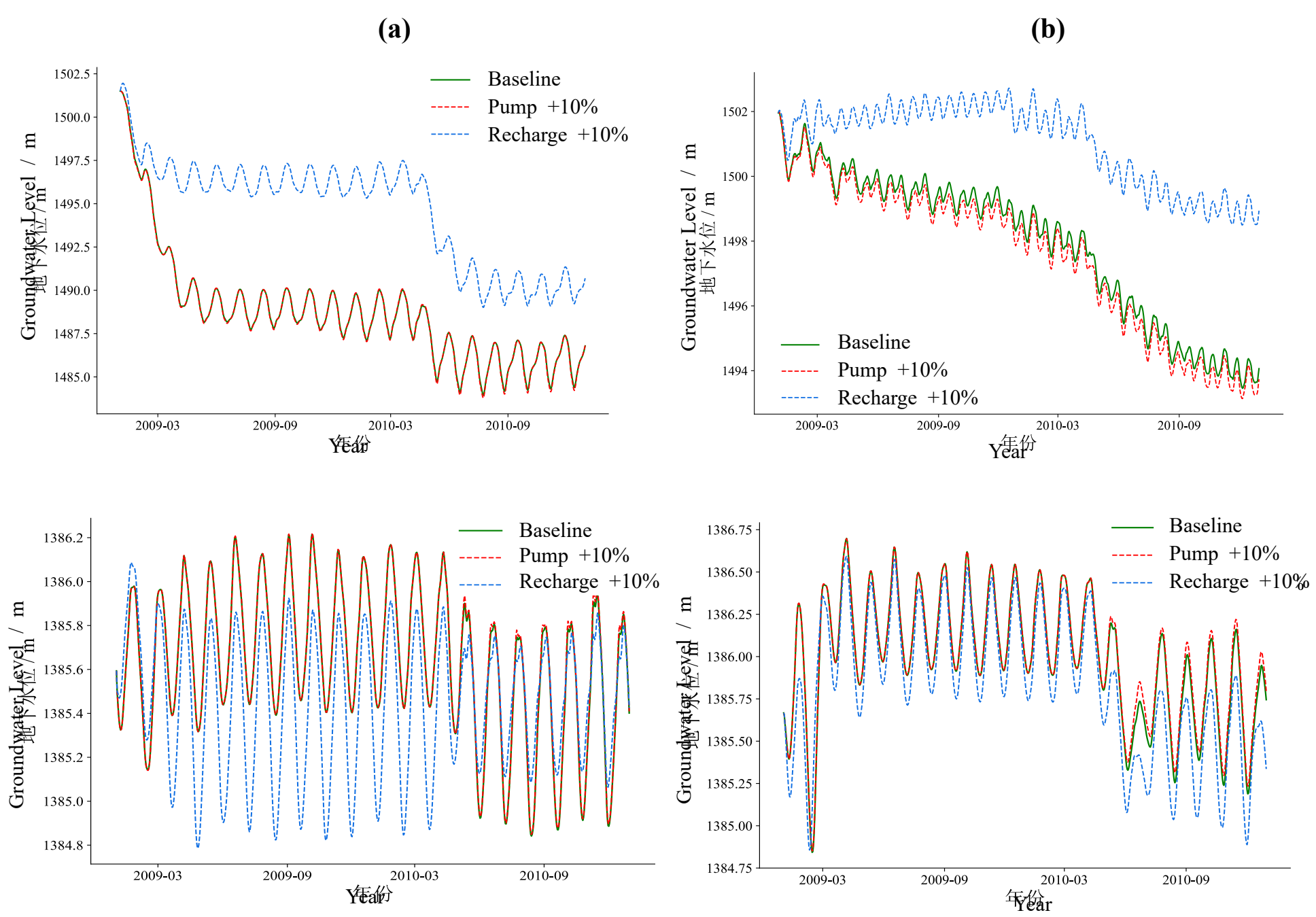

**Fig. 7 Comparison of prediction results under three scenarios of observation wells No.2 (upper) and No.28 (lower) for (a) LSTM model; and (b) Transformer model.**

The statistical quantile method is adopted to determine the historical groundwater level threshold. Specifically, the 10th percentile (P10) of groundwater level series during model training is defined as the control water level. A low-water-level risk is identified when the predicted groundwater level falls below this threshold. The thresholds are calculated as 1503.54 m for Observation Well No.2 and 1385.14 m for Observation Well No.28. The allowable duration of water levels below the threshold is set as the constraint. Groundwater levels in the piedmont inclined zone show a persistent declining trend, while those in the river valley plain fluctuate frequently. To accommodate low water levels during dry seasons or periods of insufficient recharge, the allowable duration is set to 180 days. We set the variation multipliers of pumping volume and recharge depth to -20%, -10%, 10% and 20%. Using the recursive mode, one variable is kept constant while the other is adjusted gradually, so as to determine the safe operation limits of the groundwater system under diverse pumping and recharge scenarios.

Table 9 presents the minimum water level risk indicators under different pumping scenarios. For Well No.2, the predicted water levels from both models stay below the threshold throughout the period, so the maximum allowable pumping multiplier cannot be determined. This means the groundwater level has long remained below the safety limit, with an irreversible and high risk of over-extraction. For Well No.28, both models yield a maximum allowable pumping multiplier of 1.2. Table 10 shows the minimum water level risk results under varying recharge depths. The two models exhibit distinctly different responses compared with the pumping scenarios. For Well No.2, the LSTM model captures water level recovery and risk reduction caused by increased recharge, while the Transformer focuses on the long-term declining trend and fails to reflect short-term improvements from enhanced recharge. For Well No.28, the risk of the LSTM model rises linearly as recharge decreases, with a minimum recharge multiplier of 0.9. By contrast, the Transformer shows a milder risk growth rate and is more sensitive to water level recovery driven by recharge, with its minimum recharge multiplier reaching 1.1.

**Table 9 Statistics of risk indicators below minimum water level under different pumping**

| Observation Well | Scenario Type | Model | Days Below the Treshold | Percentage of Period | Minimum Water Level (m) |
|---|---|---|---|---|---|
| 2 | -20% | LSTM | 730 | 100% | 1489.21 |
| | | Transformer | | | 1494.28 |
| | -10% | LSTM | | | 1487.56 |
| | | Transformer | | | 1493.84 |
| | 10% | LSTM | | | 1496.20 |
| | | Transformer | | | 1493.14 |
| | 20% | LSTM | | | 1486.51 |
| | | Transformer | | | 1492.86 |
| 28 | -20% | LSTM | 186 | 25.48% | 1383.98 |
| | | Transformer | 80 | 10.96% | 1384.79 |
| | -10% | LSTM | 182 | 24.93% | 1384.25 |
| | | Transformer | 80 | 10.96% | 1384.87 |
| | 10% | LSTM | 167 | 24.24% | 1384.73 |
| | | Transformer | 111 | 15.07% | 1384.88 |
| | 20% | LSTM | 156 | 24.11% | 1384.98 |
| | | Transformer | 123 | 20.15% | 1384.87 |

**Table 10 Statistics of risk indicators below minimum water level under different recharge depth**

| Observation Well | Scenario Type | Model | Days Below the Treshold | Percentage of Period | Minimum Water Level (m) |
|---|---|---|---|---|---|
| 2 | -20% | LSTM | 730 | 100% | 1496.93 |
| | | Transformer | | | 1484.72 |
| | -10% | LSTM | | | 1498.67 |
| | | Transformer | | | 1488.34 |
| | 10% | LSTM | 717 | 98.22% | 1500.41 |
| | | Transformer | 730 | 100% | 1498.50 |
| | 20% | LSTM | 449 | 61.51% | 1500.72 |
| | | Transformer | 730 | 100% | 1500.76 |
| 28 | -20% | LSTM | 0 | 0% | 1385.20 |
| | | Transformer | | | 1385.42 |
| | -10% | LSTM | 49 | 6.71% | 1385.04 |
| | | Transformer | 0 | 0% | 1385.38 |
| | 10% | LSTM | 272 | 37.26% | 1384.60 |
| | | Transformer | 153 | 23.26% | 1384.87 |
| | 20% | LSTM | 326 | 44.66% | 1384.51 |
| | | Transformer | 122 | 19.71% | 1384.83 |

In summary, the piedmont inclined plain faces structural risks from persistent water level decline. Reduced pumping and increased recharge bring limited improvement, and long-term over-extraction is difficult to reverse. The river plain groundwater system has higher resilience, requiring strict pumping control and artificial recharge to maintain the balance between groundwater and surface water. LSTM responds linearly to driving factors and judges complex trends conservatively. By contrast, Transformer accurately captures nonlinear effects and prudently evaluates high-risk conditions at Well No. 2, offering differentiated support for hydrological management decisions.

# 6 Conclusions

Considering the invisibility, complex monitoring and multi-agent governance of groundwater systems, this paper establishes a Responsible AI framework involving transparency, privacy governance, technical robustness, fairness, accountability and sustainability, and clarifies the responsibilities of all stakeholders. Taking the middle reaches of the Heihe River Basin as the study area, we compared the groundwater level prediction performance of LSTM and Transformer based on hydrometeorological data. The main results are as follows: (1) In terms of interpretability, Transformer achieves more stable feature attribution with less noise; LSTM relies on recent temporal information, while Transformer is superior in capturing long-range dependencies; (2)

Transformer has lower prediction uncertainty, stronger anti-disturbance capacity and better overall robustness; (3) Zoned groundwater regulation thresholds are proposed. For the river plain area, Transformer gives conservative predictions requiring recharge no less than 10%, while LSTM suggests recharge not exceeding 10%, and both limit the increase of pumping volume to within 20%. Both models also call for stricter pumping control and recharge measures in the piedmont inclined zone. This study verifies the good practicability of the proposed Responsible AI framework. The models and methods can provide technical and practical references for sustainable groundwater management and standardized AI application in hydrology.

## References

[1] KLEPPE A, SKREDE O J, DE RAEDT S, et al. Designing deep learning studies in cancer diagnostics[J]. Nature Reviews Cancer, 2021, 21(3): 199-211.

[2] CHOUNG H, DAVID P, LING T W. Acceptance of AI-powered facial recognition technology in surveillance scenarios: Role of trust, security, and privacy perceptions[J]. Technology in society, 2024, 79: 102721.

[3] DIGNUM V. Responsible artificial intelligence: How to develop and use AI in a responsible way[M]. Cham: Springer, 2019.

[4] SMUHA N A. The EU approach to ethics guidelines for trustworthy artificial intelligence[J]. Computer Law Review International, 2019, 20(4): 97-106.

[5] Kusche I. Possible harms of artificial intelligence and the EU AI act: fundamental rights and risk[J]. Journal of Risk Research, 2024: 1-14.

[6] POURSAEID M, POURSAEID A H, SHABANLOU S. A comparative study of artificial intelligence models and a statistical method for groundwater level prediction[J]. Water Resources Management, 2022, 36(5): 1499-1519.

[7] RICHARDS C E, TZACHOR A, AVIN S, et al. Rewards, risks and responsible deployment of artificial intelligence in water systems[J]. Nature Water, 2023, 1(5): 422-432.

[8] LI B, QI P, LIU B, et al. Trustworthy AI: From principles to practices[J]. ACM Computing Surveys, 2023, 55(9): 1-46.

[9] HANNA R, KAZIM E. Philosophical foundations for digital ethics and AI Ethics: a dignitarian approach[J]. AI and Ethics, 2021, 1(4): 405-423.

[10] DIAZ-RODRIGUEZ N, DEL SER J, COECKELBERGH M, et al. Connecting the dots in trustworthy Artificial Intelligence: From AI principles, ethics, and key requirements to responsible AI systems and regulation[J]. Information Fusion, 2023, 99: 101896.

[11] HOCHREITER S, SCHMIDHUBER J. Long short-term memory[J]. Neural Computation, 1997, 9(8): 1735-1780.
[12] VASWANI A, SHAZEER N, PARMAR N, et al. Attention is all you need[C]// Proceedings of the 31st International Conference on Neural Information Processing Systems. Long Beach, CA, USA: Curran Associates, Inc, 2017: 5998-6008.
[13] ARRIETA A B, DIAZ-RODRIGUEZ N, DEL SER J, et al. Explainable Artificial Intelligence (XAI): Concepts, taxonomies, opportunities and challenges toward responsible AI[J]. Information fusion, 2020, 58: 82-115.
[14] KOZIELSKI M, SIKORA M, WAWROWSKI L. Towards consistency of rule-based explainer and black box model-Fusion of rule induction and XAI-based feature importance[J]. Knowledge-Based Systems, 2025, 311: 113092-113101.
[15] XIE Z, LIU L, WANG C. Model-guided boosting for image denoising[J]. Signal Processing, 2022, 201: 108721.
[16] TSOUKALAS D, MITTAS N, ARVANITOU E M, et al. Local and global explainability for technical debt identification[J]. IEEE Transactions on Software Engineering, 2024, 50(8): 2110-2123.
[17] PAPAGIANNIDIS E, MIKALEF P, CONBOY K. Responsible artificial intelligence governance: A review and research framework[J]. The Journal of Strategic Information Systems, 2025, 34(2): 101885-101903.
[18] GUNDERSEN O E. The fundamental principles of reproducibility[J]. Philosophical Transactions of the Royal Society A, 2021, 379(2197): 20200210.
[19] KLEANTHOUS S, IOANNOU A, PAPANDREOU F. Social Influence and the Perceived Fairness of Algorithmic Decisions: An Exploratory Study[C]// Adjunct Proceedings of the 33rd ACM Conference on User Modeling, Adaptation and Personalization. 2025: 87-92.